\documentclass{tlp}

\usepackage{amsmath}
\usepackage{graphicx}
\usepackage{multirow}

\usepackage{soul}
\usepackage{url}
\usepackage[hidelinks]{hyperref}
\usepackage[utf8]{inputenc}
\usepackage{graphicx}
\usepackage{subfigure}
\usepackage{amssymb}
\usepackage{booktabs}
\usepackage{algorithm}
\usepackage{algorithmic}
\usepackage{xspace}

\newcommand\code[1]{\texttt{#1}}

\newcommand\KALMF{{KALM\textsuperscript{FL}}\xspace}
\newcommand\KALMR{{KALM\textsuperscript{RA}}\xspace}
\newcommand\Stanza{{\textsc{Stanza}}\xspace}
\newcommand\MS{{m\textsc{Stanza}}\xspace}

\newcounter{example}
\newenvironment{example}{\refstepcounter{example}\noindent\textbf{Example~\theexample}}

\newcounter{property}
\newcounter{definition}
\newenvironment{definition}{\refstepcounter{definition}\noindent\textbf{Definition~\thedefinition}}

\begin{document}

\lefttitle{Yuheng Wang, Paul Fodor and Michael Kifer}

\jnlPage{1}{8}
\jnlDoiYr{2021}
\doival{10.1017/xxxxx}

\title[Knowledge Authoring for Rules and Actions]{Knowledge Authoring for Rules and Actions\thanks{Research partially funded by NSF grant 1814457.}}

\begin{authgrp}
\author{\sn{Yuheng} \gn{Wang}, \sn{Paul} \gn{Fodor}, and \sn{Michael} \gn{Kifer}}
\affiliation{Stony Brook University, Stony Brook, NY, USA}
\email{\{yuhewang,pfodor,kifer\}@cs.stonybrook.edu}{}
\end{authgrp}

\history{\sub{xx xx xxxx;} \rev{xx xx xxxx;} \acc{xx xx xxxx}}

\maketitle

\begin{abstract}
Knowledge representation and reasoning (KRR) systems describe and reason with complex concepts and relations in the form of facts and rules.
Unfortunately, wide deployment of KRR systems runs into the problem that domain experts have great difficulty constructing correct logical representations of their domain knowledge.
Knowledge engineers can help with this construction process, but there is a deficit of such specialists.
The earlier Knowledge Authoring Logic Machine (KALM, \cite{gao2018knowledge}) based on Controlled Natural Language (CNL) was shown to have very high accuracy for authoring facts and questions.
More recently, \KALMF (\cite{wang2022knowledge}), a successor of KALM, replaced CNL with \textit{factual} English, which is much less restrictive and requires very little training from users.
However, \KALMF has limitations in representing certain types of knowledge, such as authoring rules for multi-step reasoning or understanding actions with timestamps.
To address these limitations, we propose \KALMR to enable authoring of rules and actions.
Our evaluation using the UTI guidelines benchmark shows that \KALMR achieves a high level of correctness (100\%) on rule authoring. 
When used for authoring and reasoning with actions, \KALMR achieves more than 99.3\% correctness on the bAbI benchmark, demonstrating its effectiveness in more sophisticated KRR jobs. 
Finally, we illustrate the logical reasoning capabilities of \KALMR by drawing attention to the problems faced by the recently made famous AI, ChatGPT.
\end{abstract}



\setcounter{tocdepth}{5}


\section{Introduction}

Knowledge representation and reasoning (KRR) systems represent human knowledge as facts, rules, and other logical forms.
However, transformation of human knowledge to these logical forms requires the expertise of knowledge engineers with KRR skills, which,
unfortunately, is scarce.

To address the shortage of knowledge engineers, researchers have explored the use of different languages and translators for representing human knowledge.
One idea was to use natural language (NL), but the NL-based systems, such as OpenSesame (\cite{swayamdipta2017frame}) and SLING (\cite{ringgaard2017sling}), had low accuracy, and led to significant errors in subsequent reasoning.
The accuracy issue then motivated researchers to consider Controlled Natural Language (CNL, \cite{fuchs2008attempto,schwitter2002english}) for knowledge authoring. 
Unfortunately, although
CNL does improve accuracy, it is hard for a typical user (say, a domain expert) to learn
a CNL grammar and its syntactic restrictions.
Furthermore, systems based on either NL or CNL cannot identify sentences with the same meaning but different forms. For example, ``\textit{Mary buys a car}" and ``\textit{Mary makes a purchase of a car}" would be translated into totally different logical representations. This problem, known as \textit{semantic mismatch} (\cite{gao2018high}), is a serious limitation affecting accuracy.

The Knowledge Authoring Logic Machine (KALM, \cite{gao2018knowledge}) was introduced to tackle semantic mismatch problem, but this approach was based on a CNL (Attempto, \cite{fuchs2008attempto})
and had heavy syntactic limitations.
Recently, the \KALMF system (\cite{wang2022knowledge}) greatly relaxed these restrictions by focusing on \emph{factual} English sentences, which are suitable for expressing facts and queries and require little training to use.
To parse factual sentences, \KALMF replaced the CNL parser in the original KALM system with an improved neural NL parser called \MS.
However, this alteration brought about several new issues that are typical in neural parsers, such as errors in part-of-speech and dependency parsing.
\KALMF then effectively addressed these issues and achieved high accuracy in  authoring facts and queries with factual sentences.

In this paper, we focus on other types of human knowledge that \KALMF does not cover, such as, rules and actions.
We further extend \KALMF to support authoring of rules and actions, creating a new system called KALM for Rules and Actions (or \KALMR).\footnote{
\url{https://github.com/yuhengwang1/kalm-ra}
}
\KALMR allows users to author rules using factual sentences and perform multi-step frame-based reasoning using F-logic (\cite{kifer1989f}).
In addition to rule authoring, \KALMR incorporates a formalism known as Simplified Event Calculus (SEC, \cite{sadri1995variants}) to represent and reason about actions and their effects.
The use of authored knowledge (facts, queries, rules, and actions) allows for logical reasoning within an \textit{underlying logical system for reasoning with the generated knowledge}.
This system must align with the scope of the knowledge that \KALMR can represent, and supports the inference of new knowledge from existing one.
In terms of implementation, we found a Prolog-like system is more suitable for frame-based parsing, so we implemented \KALMR in XSB (\cite{swift2012xsb}).
However, the knowledge produced by \KALMR contains disjunctive knowledge and function symbols, so we chose the answer set programming system DLV (\cite{leone2006dlv}) as the logical system for reasoning about the generated knowledge.\footnote{
    Other ASP logic programming systems, such as Potassco (\cite{gebser2019multi}), lack the necessary level of support for function symbols and querying.}
Evaluation on benchmarks including the UTI guidelines (\cite{shiffman2009writing}) and bAbI Tasks (\cite{weston2015towards}) shows that \KALMR achieves 100\% accuracy on authoring and reasoning with rules, and 99.3\% on authoring and reasoning about actions.
Finally, we assess the recently released powerful dialogue model,
ChatGPT\footnote{\url{https://chat.openai.com/chat}}, using bAbI Tasks,
and highlight its limitations with respect to logical reasoning compared to \KALMR. 

The paper is organized as follows: Section \ref{sec:related} reviews the \KALMF system and some logic programming techniques, Section \ref{sec:kalmr} introduces the new \KALMR system and describes how it represents rules and actions, Section \ref{sec:eval} presents the evaluation settings and results, and Section \ref{sec:conclusion} concludes the paper and discusses future work.



\section{Background}\label{sec:related}

\subsection{Knowledge Authoring Logic Machine for Factual Language}
The Knowledge Authoring Logic Machine (KALM, \cite{gao2018knowledge,gao2018high}) allows users to author knowledge using Attempto Controlled English (ACE, \cite{fuchs1996attempto}).
However, ACE's grammar is too limiting and poses a high learning curve, particularly for non-technical users.
To mitigate this problem, KALM was extended to KALM for Factual (English) Language (\KALMF, \cite{wang2022knowledge}) by introducing \emph{factual (English) sentences} and focusing on authoring facts and simple queries.
Factual sentences express atomic database facts and queries (e.g. ``\textit{Mary buys a car}").
They can become more complex with adnominal clauses (e.g. ``\textit{Mary buys a car that is old}") and can be combined via ``\textit{and}" and ``\textit{or}" (e.g. ``\textit{Mary buys a car and Bob buys a watch}").
In comparison, sentences not expressing factual information (e.g. ``\textit{Fetch the ball}" or ``\emph{Oh, well}") are non-factual and are not allowed.
Factual sentences can be captured through properties based on dependency analysis and Part-of-Speech (POS) tagging (\cite{wang2022knowledge}), which is a very mild restriction compared to complex grammars such as in ACE.
This means that users do not need to master complex grammars.
Instead, they can simply write normal sentences that describe database facts, or basic Boolean combinations of facts and, as long as they avoid fancy language forms, their sentences will be accepted.

\begin{figure}[htbp!]
     \centering
     \subfigure[Training stage]{\label{fig:training}\includegraphics[width=110mm]{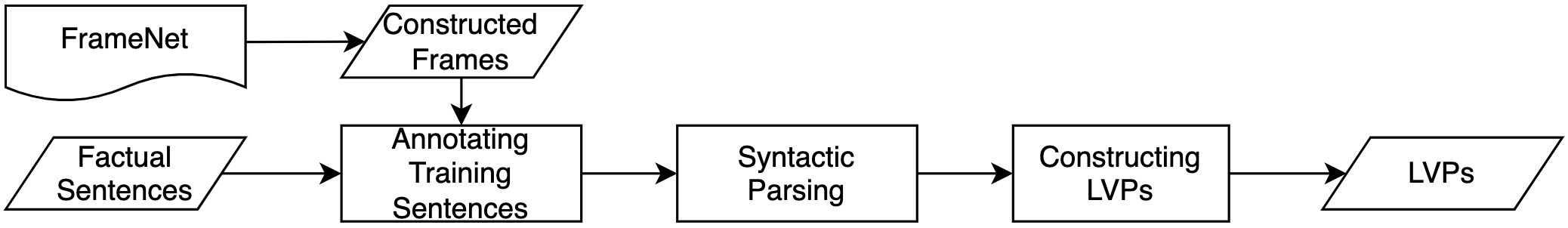}}
     \par\vspace{-2mm}
     \subfigure[Deployment stage]{\label{fig:parsing}\includegraphics[width=110mm]{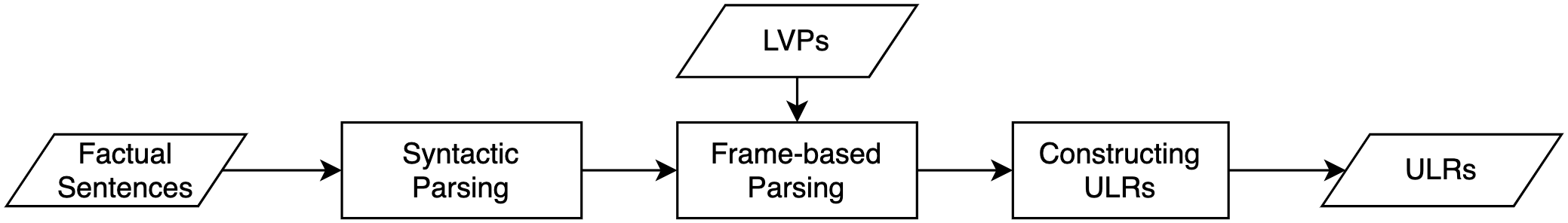}}
     \vspace{-1mm}
     \caption{The frameworks of the \KALMF system}
     \label{fig:framework}
\end{figure}

\KALMF is a two-stage system following the \emph{structured machine learning} paradigm.
In the first stage, known as the \textit{training stage}, \KALMF constructs \emph{logical valence patterns} (LVPs) by learning from \textit{training sentences}. An LVP is
a specification that tells how to extract \emph{role fillers} for the concepts represented by the English sentences
related to that LVP.
In the second stage, known as the \textit{deployment stage}, the system does semantic parsing by applying the constructed LVPs to convert factual English sentences into \emph{unique logical representations} (ULRs).
Fig.~\ref{fig:training} depicts the training stage of \KALMF, with the key steps explained in the accompanying text.

\noindent
\textbf{Annotating Training Sentences.}
To enable semantic understanding of a domain of discourse,
knowledge engineers must first construct the required background knowledge in the form of \KALMF frames. The overall structure of most of these frames can be adopted from FrameNet (\cite{baker1998berkeley}) and converted into the logic form required by \KALMF.
Then, knowledge engineers compose training sentences and annotate them using \KALMF frames.
For example, the annotated training sentence~(\ref{code:training}), below,  indicates that 
the meaning of ``\textit{Mary buys a car}" is captured by the \code{Commerce\_buy} frame;
the word that triggers this frame, a.k.a. the \emph{lexical unit} (LU), is the 2nd word ``\textit{buy}" or its synonym ``\textit{purchase}"; and,
the 1st and the 4th words, ``\textit{mary}" and ``\textit{car}", play the roles of \code{Buyer} and \code{Goods} in the frame.

\vspace{-1mm}
{\small
\begin{equation}
\label{code:training}
\hspace*{-40mm}
\begin{aligned}
\verb|train("Mary buys a car","Commerce_buy","LU"=2,[|&\verb|purchase],|\\
\verb|["Buyer"=1+required,"Goods"=4+required]).|
\end{aligned}
\vspace{1mm}
\end{equation}}

\begin{figure}[htbp!]
     \centering
     \subfigure[]{\label{fig:ms1}\includegraphics[width=45mm]{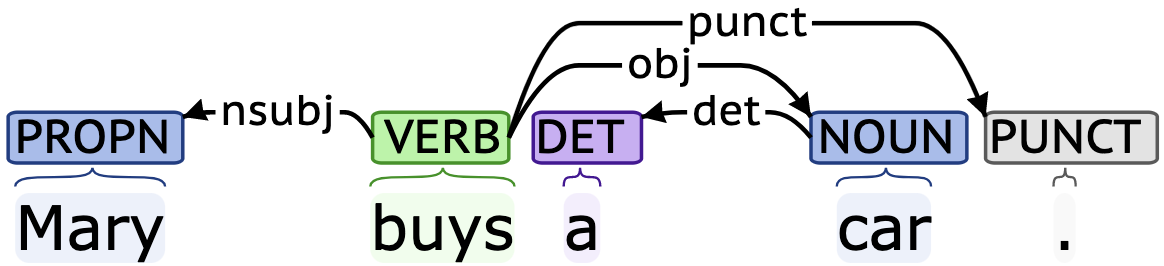}}
     \hspace{5mm}
     \subfigure[]{\label{fig:ms2}\includegraphics[width=45mm]{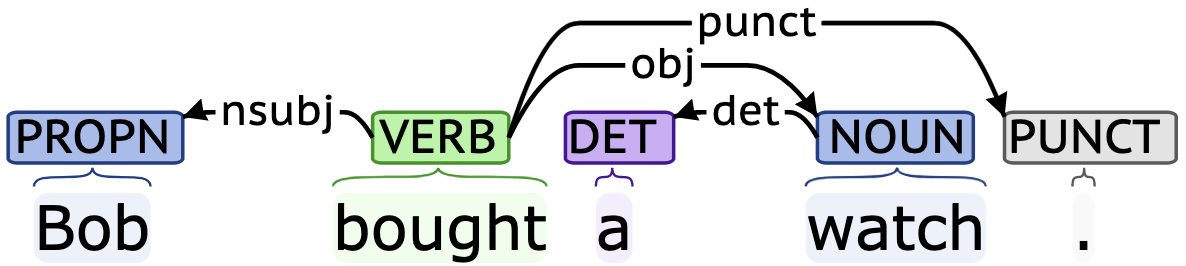}}
     \vspace{-2mm}
     \caption{\MS parses}
     \label{fig:ms}
\end{figure}

\noindent
\textbf{Syntactic Parsing.}
\KALMF then performs syntactic parsing using \MS (\cite{wang2022knowledge})\footnote{
A modification of \Stanza (\cite{qi2020stanza}) that returns ranked lists of parses rather than just one parse.
}
and automatically corrects some parsing errors.
Fig.~\ref{fig:ms} shows two \MS parses, where the colored boxes contain POS tags and the labeled arrows display dependency relations.

\vspace{-1mm}

\vspace{-3mm}
\noindent
\textbf{Construction of LVPs.}
\MS parses, along with annotations of sentences, allow \KALMF to construct LVPs that specify how to fill the roles of a frame triggered by an LU.
For example, by synthesizing the information in training sentence~(\ref{code:training}) and the \MS parse in Fig.~\ref{fig:ms1}, \KALMF learns that, to fill the roles \code{Buyer} and \code{Goods} of the \code{Commerce\_buy} frame triggered by the LU ``\textit{buys}",
one should extract the subject and object of ``\textit{buys}" through the dependency relations \code{nsubj} and \code{obj}, respectively.
This learned knowledge about role-filling is encoded as an LVP~(\ref{code:lvp}) as follows:

\vspace{-1mm}
{\small
\begin{equation}\label{code:lvp}
\hspace*{-39mm}
\begin{aligned}
\verb|lvp(buy,"Commerce_buy",[pattern("Buyer",[nsubj],required),|\\
\verb|pattern("Goods",[obj],required)]).|
\end{aligned}
\vspace{1mm}
\end{equation}}

The deployment stage of \KALMF is illustrated in Fig.~\ref{fig:parsing}.
Two key steps in this stage are further explained below.

\noindent
\textbf{Frame-based Parsing.}
When an unseen factual sentence comes in, \KALMF triggers all possible LVPs using the words in the sentence.
Then, the triggered LVPs are applied to this sentence to extract role fillers and a frame-based parse of this sentence is generated.
For example, a new sentence ``\textit{Bob bought a watch}" is parsed as Fig.~\ref{fig:ms2} by \MS. It triggers LVP~(\ref{code:lvp}) using the LU ``\textit{bought}" (whose base form is ``\textit{buy}").
\KALMF then extracts the role filler ``\textit{Bob}" for the role \code{Buyer} according to the dependency list \code{[nsubj]}. Similarly, ``\textit{watch}" is extracted for the role \code{Goods} according to the dependency \code{[obj]}.



\noindent
\textbf{Constructing ULRs.}
Ultimately, frame-based parses are represented as ULR facts that capture the meaning of the original English sentences and are suitable for querying.
For example, the ULR for the factual sentence ``\textit{Bob
bought a watch}", below, indicates that the meaning of the sentence is captured by the \code{Commerce\_buy} frame,
that ``\textit{Bob}" is the \code{Buyer}, and ``\textit{watch}" plays the role of \code{Goods},
where \code{rl/2} represents instances of (role, role-filler) pairs.

{\small
\begin{verbatim}
frame("Commerce_buy",[rl("Buyer","Bob"),rl("Goods",watch)]).
\end{verbatim}}




\subsection{Disjunctive Information and Frame Reasoning}\label{sec:lp}

Our reasoning subsystem combines Answer Set Programming (ASP) with aspects of frame-based reasoning.

DLV (\cite{leone2006dlv}) is a disjunctive version of Datalog that operates under the ASP paradigm.
It extends Datalog by adding support for disjunction in facts and rule heads, thus providing greater expressiveness for disjunctive information than KRR systems based on the well-founded semantics (e.g., XSB, \cite{swift2012xsb}).
Furthermore, DLV's support for function symbols and querying makes it more convenient for working with frames (\cite{fillmore2006frame}) than other ASP systems, such as Potassco (\cite{gebser2019multi}).

F-logic (\cite{flogic-95,kifer1989f}) is a knowledge representation and ontology language that combines the benefits of conceptual modeling with object-oriented and frame-based languages.
One of its key features is the ability to use \textit{composite frames} to reduce long conjunctions of roles into more compact forms, matching ideally the structure of FrameBase's frames.
For example, F-logic frames\footnote{
    We depart from the actual syntax of F-logic as it is not supported by the DLV system.
    Instead, we implemented a small subset of that logic by casting it directly into the already supported DLV syntax.
}
can be used to answer the question ``\textit{What did Mary buy?}" given the fact ``\textit{Mary bought a car for Bob}," whose ULRs, shown below, are not logically equivalent (the fact has more roles than the query).

{\small
\begin{verbatim}
frame("Commerce_buy",[rl("Buyer","Mary"),rl("Goods",car),rl("Recipient","Bob")]).
?- frame("Commerce_buy",[rl("Buyer","Mary"),rl("Goods",What)]).   What=car.
\end{verbatim}}


\subsection{Event Calculus for Reasoning about Actions and their Effects}\label{sec:ec}
The event calculus (EC, \cite{kowalski1989logic}) is a set of logical axioms that describe the \textit{law of inertia} for actions. This law states that time-dependent facts, 
 \textit{fluents}, that are not explicitly changed by an action preserve their true/false status in the state produced by that action.
 Here we use 
the simplified event calculus (SEC, \cite{sadri1995variants}),
which is a simpler and more tractable variant of the original EC.
A fluent in SEC is said to hold at a particular timestamp if it is initiated by an action and not terminated subsequently.
This is formalized by these DLV rules:

{\small
\begin{verbatim}
holdsAt(F,T2) :-
    happensAt(A,T1), initiates(A,F), timestamp(T2), T1 < T2,
    not stoppedIn(T1,F,T2).
stoppedIn(T1,F,T2) :- 
    happensAt(A,T), terminates(A,F), timestamp(T1), T1 < T, timestamp(T2), T < T2.
\end{verbatim}}

Here \code{happensAt/2} represents a momentary occurrence of action \code{A} at a timestamp.
If an action is exogenous insertion of a fluent $f$ at time $t$ then we also represent it as \code{happensAt(f,t)}.
Example \ref{exmp:13} demonstrates the use of \code{happensAt/2}.

\begin{example}
\label{exmp:13}
The sentence ``\textit{Mary goes to the bedroom. The bedroom is north of the garden.}" is represented as follows:

{\small
\begin{verbatim}
happensAt(frame("Travel",[rl("Person","Mary"),rl("Place",bedroom)]),1).
happensAt(frame("North_of",[rl("Entity1",bedroom),rl("Entity2",garden)]),2).
person("Mary"). place(bedroom). entity(bedroom). entity(garden). timestamp(1..2).
\end{verbatim}}
\noindent
The first \code{happensAt/2} introduces an action of traveling from place to place while the second \code{happensAt/2} uses an observed (i.e., exogenously inserted) fluent \code{"North\_of"(bedroom,garden)}.
Observable fluents are supposed to be disjoint from action fluents, and we will use a special predicate, \texttt{observable/1}, to recognize them in SEC rules.
Timestamps indicate the temporal relation between the action and the observed fluent.
Predicates \code{person/1}, \code{place/2}, \code{entity/2}, define the domain of roles, while \code{timestamp/1} restricts the domain of timestamps.
\end{example}


The predicates
\code{initiates(Action,Fluent)} and \code{terminates(Action,Fluent)} in SEC are typically used to specify domain-specific axioms that capture the initiation and termination of fluents.

\vspace{-4mm}
\section{Extending \texorpdfstring\KALMF{} for Rules and Actions}\label{sec:kalmr}
This section describes an extension of \KALMF to handle rules and actions (\KALMR).

Since we want to be able to handle disjunctive information required by some of the bAbI tasks, we made a decision to switch the reasoner from XSB which was used in \KALMF to an ASP-based system DLV (\cite{leone2006dlv}) that can handle disjunction in the rule heads.
Thus, the syntax of the ULR, i.e., the logical statements produced by \KALMR, follows that of DLV.
A number of examples inspired by the UTI guidelines and bAbI Tasks are used in this section to illustrate the workings of \KALMR.

\subsection{Authoring and Reasoning with \texorpdfstring\KALMR{} Rules}\label{sec:ruleauth}
Rules are important to KRR systems because they enable multi-step logical inferences needed for real-world tasks, such as diagnosis, planning, and decision making.
Here we address the problem of rule authoring.

\subsubsection{Enhancements for Representation of Facts}\label{sec:factrep}

First we discuss the representation of disjunction, conjunction, negation, and coreference, which is not covered in \KALMF.

\textbf{Conjunction and Disjunction.}
The \KALMR system prohibits the use of a mixture of conjunction and disjunction within a single factual sentence to prevent ambiguous expressions such as ``\textit{Mary wants to have a sandwich or a salad and a drink.}"
To represent a factual sentence with homogeneous conjunction or disjunction,
the system first parses the sentence into a set of component ULRs.
For conjunction, 
\KALMR uses this set of ULRs as the final representation.
For disjunction, the component ULRs are assembled into a single disjunctive ULR using DLV's disjunction \code{v} as shown in
Example \ref{exmp:2}.

\begin{example}
\label{exmp:2}
The factual sentence with conjunction ``\textit{Daniel administers a parenteral and an oral antimicrobial therapy for Mary}'' is represented as the following set of ULRs:

{\small
\begin{verbatim}
frame("Cure",[rl("Doctor","Daniel"),rl("Patient","Mary"),
              rl("Therapy",antimicrobial),rl("Method",parenteral)]).
frame("Cure",[rl("Doctor","Daniel"),rl("Patient","Mary"),
                    rl("Therapy",antimicrobial),rl("Method",oral)]).
doctor("Daniel"). patient("Mary"). therapy(antimicrobial).
method(parenteral). method(oral).
\end{verbatim}}

\noindent
where the predicates \code{doctor}, \code{patient}, \code{therapy}, and \code{method} define the domains for the roles.
These domain predicates will be omitted in the rest of the paper, for brevity.

The disjunctive factual sentence 
``\textit{Daniel administers a parenteral or an oral antimicrobial therapy for Mary}" is represented as the following ULR:

{\small
\begin{verbatim}
frame("Cure",[rl("Doctor","Daniel"),rl("Patient","Mary"),
              rl("Therapy",antimicrobial),rl("Route",parenteral)])
v frame("Cure",[rl("Doctor","Daniel"),rl("Patient","Mary"),
                rl("Therapy",antimicrobial),rl("Route",oral)]).
\end{verbatim}}
\end{example}

\textbf{Negation.}\label{sec:neg}
The \KALMR system supports \emph{explicit negation} through the use of the negative words ``\textit{not}" and ``\textit{no}".
Such sentences are captured by appending the suffix ``\code{\_not}" to the name of the frame triggered by this sentence.

\begin{example}
\label{exmp:3}
The explicitly negated factual sentence ``\textit{Daniel's patient Mary does not have UTI}'' is represented by

{\small
\begin{verbatim}
frame("Medical_issue_not",[rl("Doctor","Daniel"),rl("Patient","Mary"),
                           rl("Ailment","UTI")]).
\end{verbatim}}
\end{example}
\noindent

\textbf{Coreference.}
Coreference occurs when a word or a phrase refers to something that is mentioned earlier in the text.
Without coreference resolution, one gets unresolved references to unknown entities in ULRs.
To address this issue, \KALMR uses 
a coreference resolution tool
neuralcoref,\footnote{\url{https://github.com/huggingface/neuralcoref}}
which
identifies and replaces coreferences with the corresponding entities from the preceding text.

\begin{example}
\label{exmp:4}
The factual sentences ``\textit{Daniel's patient Mary has UTI. He administers an antimicrobial therapy for her.}"
are turned into

{\small
\begin{verbatim}
frame("Medical_issue",[rl("Doctor","Daniel"),rl("Patient","Mary"),
                       rl("Ailment","UTI")]).
frame("Cure",[rl("Doctor","Daniel"),rl("Patient","Mary"),
              rl("Therapy",antimicrobial)]).
\end{verbatim}}
\end{example}
\noindent
where the second ULR uses entities \code{"Daniel"} and \code{"Mary"}  instead of the pronouns ``\textit{he}" and ``\textit{she}."


\subsubsection{Rule Representation}\label{sec:ruleparsing}

Rules in \KALMR are expressed in a much more restricted syntax
compared to facts since, for knowledge authoring purposes,
humans have little difficulty learning and complying with the restrictions. Moreover, since variables play such a key role in rules, complex coreferences must be specified unambiguously.
All this makes writing rules in a natural language into a very cumbersome, error-prone, and ambiguity-prone task
compared to the restricted syntax below.

\begin{definition}
\label{def:1}
A \textit{rule} in \KALMR is an if-then statement of the form ``If $P_1$, $P_2$, ..., and $P_n$, then $C_1$, $C_2$, ..., or $C_m$", where
\begin{enumerate}
    \item each $P_i$ ($i=1..n$) is a factual sentence without disjunction;
    \item each $C_j$ ($j=1..m$) is a factual sentence without conjunction;
    \item variables in $C_j$ ($j=1..m$) must use the \textit{explicitly typed syntax} (\cite{gao2018high}) and must appear in at least one of the $P_i$ ($i=1..n$). E.g., in the rule ``\textit{If Mary goes to the hospital, then \$doctor sees Mary}", the explicitly typed variable \textit{\$doctor} appears in the conclusion without appearing in the premise, which is prohibited.
    Instead, the rule author must provide some information about the doctor in a rule premise (e.g., ``and she has an appointment with \textit{\$doctor}"). This corresponds to the well-known ``rule safety" rule in  logic programming.
    \item variables that refer to the same thing must have the same name. E.g., in the rule ``\textit{If \$patient is sick, then \$patient goes to see a doctor}", the two \textit{\$patient} variables are intended to refer to the same person and thus have the same name.
\end{enumerate}
\end{definition}
\noindent

Here are some examples of rules in \KALMR.

\begin{example}
\label{exmp:6}
The \KALMR rule ``If \textit{\$doctor's \$patient is a young child and has an unexplained fever}, then \textit{\$doctor assesses \$patient's degree of toxicity or dehydration}" is represented as follows:

{\small
\begin{verbatim}
frame("Assessing",[rl("Doctor",Doctor),rl("Patient",Patient),
                   rl("Item",toxicity)])
v frame("Assessing",[rl("Doctor",Doctor),rl("Patient",Patient),
                     rl("Item",dehydration)]) :-
    frame("People_by_age",[rl("Person",Patient),rl("Type",child)]),
    frame("Medical_issues",[rl("Doctor",Doctor),rl("Patient",Patient),
                            rl("Ailment",fever),rl("Cause",unexplained)]).
\end{verbatim}}
\end{example}

\KALMR supports two types of negation in rules: \textit{explicit negation} (\cite{gelfond1991classical}) and \textit{negation as failure} (with the stable model semantics, \cite{gelfond1988stable}). 
The former allows users to specify explicitly known negative factual information while the latter lets one derive negative information from the lack of positive information.
Explicit negation in rules is handled the same way as in fact representation.
Negation as failure must be indicated by the rule author through the idiom ``\textit{not provable}", which is then converted into the predicate \code{not/1}.
The idiom ``\textit{not provable}" is prohibited in rule heads.


\begin{example}
\label{exmp:10}
The \KALMR rule ``If \textit{not provable \$doctor does not administer \$therapy for \$patient}, then \textit{\$patient undergoes \$therapy from \$doctor}"is represented as follows:

{\small
\begin{verbatim}
frame("Undergoing",[rl("Doctor",Doctor),rl("Patient",Patient),
                    rl("Therapy",Therapy)]) :-
    not frame("Cure_not",[rl("Doctor",Doctor),rl("Patient",Patient),
                          rl("Therapy",Therapy)]),
    patient(Patient), doctor(Doctor), therapy(Therapy).
\end{verbatim}}
\end{example}
\noindent
where \code{patient/1}, \code{doctor/1}, and \code{therapy/1} are domain predicates that ensure that variables that appear under negation have well-defined domains.

\subsubsection{Queries and Answers}
Queries in \KALMR must be in factual English and end with a question mark.
\KALMR translates both \textit{Wh}-variables and explicitly typed variables into the corresponding DLV variables.
Example \ref{exmp:11} shows how \KALMR represents a query with variables.

\begin{example}
\label{exmp:11}
The query ``\textit{Who undergoes \$therapy?}" has the following ULR:

{\small
\begin{verbatim}
frame("Undergoing",[rl("Patient",Who),rl("Therapy",Therapy)])?
\end{verbatim}}
\end{example}

\KALMR then invokes the DLV reasoner to compute query answers.
DLV has two inference modes: \textit{brave reasoning} and
\textit{cautious reasoning}.
In brave reasoning, a query returns answers that are true in \textit{at least one} model of the program and 
cautious reasoning returns the answers that are true in \textit{all} models. Users are free to choose either mode.

\begin{example}
\label{exmp:12}
For instance, if the underlying information contains only this single fact

{\small
\begin{verbatim}
{frame("Undergoing",[rl("Patient","Mary"),rl("Therapy",mental)]).}
\end{verbatim}}
\noindent
then there is only one model and both modes return the same result:
{\small
\begin{verbatim}
{Who="Mary",Therapy=mental}
\end{verbatim}}

\noindent
In case of
``\textit{Mary or Bob undergoes a mental therapy}", two models are are computed:

{\small
\begin{verbatim}
{frame("Undergoing",[rl("Patient","Mary"),rl("Therapy",antimicrobial)]).}
{frame("Undergoing",[rl("Patient","Bob"),rl("Therapy",antimicrobial)]).}
\end{verbatim}}
\noindent
In the cautious mode there would be no answers while the brave mode yields two:
{\small
\begin{verbatim}
{Who="Mary",Therapy=mental}
{Who="Bob",Therapy=mental}
\end{verbatim}}
\end{example}

\subsection{Authoring and Reasoning with Actions}\label{sec:actauth}
Time-independent facts and rules discussed earlier are knowledge that persists over time.
In contrast, actions are momentary occurrences of events that change the underlying knowledge, so actions are associated with timestamps.
Dealing with actions and their effects, also known as \emph{fluents}, requires an understanding of the passage of time.
\KALMR allows users to state actions using factual English and then formalizes actions as temporal database facts using SEC discussed Section \ref{sec:ec}.
The following discussion of authoring and reasoning with actions will be in the SEC framework.
Reasoning based on SEC requires the knowledge of fluent initiation and termination.
This information is part of the commonsense and domain knowledge supplied by knowledge engineers and domain experts via high-level fluent initiation and termination statements (Definition \ref{def:2}) and \KALMR translates them into facts and rules that involve the predicates \code{initiates/2} and \code{terminates/2} used by Event Calculus. 
Knowledge engineers supply the commonsense part of these statements and domain experts supply the domain-specific part. 

\begin{definition}\label{def:2}
    A \emph{fluent initiation statement} in \KALMR has the form ``\textit{$A$/$F_{obs}$ initiates $F_{init}$}" and a \emph{fluent termination statement} in \KALMR has the form ``\textit{$A$/$F_{obs}$ terminates $F_{term}$}", where

\begin{enumerate}
    \item action $A$, observed fluent $F_{obs}$, initiated fluent $F_{init}$, and terminated fluent $F_{term}$ are factual sentences without conjunction or disjunction;
    \item variables in $F_{init}$ use explicitly typed syntax and must appear in $A$ (or in $F_{obs}$ when a fluent is observed) to avoid unbound variables in initiated fluents;
    \item variables that refer to the same thing must have the same name.
\end{enumerate}
\end{definition}

Example \ref{exmp:5} shows how \KALMR represents fluent initiation and termination.

\begin{example}
\label{exmp:5}
The commonsense initiation statement ``\textit{\$person travels to \$place initiates \$person is located in \$place}" would be created by a knowledge engineer and translated by \KALMR as the following rule:

{\small
\begin{verbatim}
initiates(frame("Travel",[rl("Person",Person),rl("Place",Place)]),
          frame("Located",[rl("Entity",Person),rl("Location",Place)])):-
    person(Person), place(Place).
\end{verbatim}}
\noindent
Here \code{person/1} and \code{place/1} are used to guarantee rule safety.
Since, any object can be in one place only at any given time, we have a commonsense termination statement ``\textit{\$person travels to \$place1 terminates \$person is located in \$place2.}"
This statement would also be created by knowledge engineers
and translated by \KALMR as follows:

{\small
\begin{verbatim}
terminates(frame("Travel",[rl("Person",Person),rl("Place",Place)]),
           frame("Located",[rl("Entity",Person),rl("Location",Place2)])):-
    person(Person), place(Place), entity(Person), location(Place2), Place!=Place2.
\end{verbatim}}
\noindent
\end{example}

\KALMR also enhances rules by incorporating temporal information, allowing the inference of new knowledge under the SEC framework.
The process begins by requiring users to specify their domain knowledge on fluents in the form of rules described in Definition \ref{def:1}.
Then \KALMR translates these rules into ULRs, with each premise and conclusion linked to a timestamp via the \code{holdsAt/2} predicate.
We call these rules \emph{time-related} because they enable reasoning with fluents containing temporal information.
Here is an example of a time-related rule.

{\small
\begin{verbatim}
holdsAt(ULRC1,T) v ... v holdsAt(ULRCm,T) :-
    holdsAt(ULRP1,T), ..., holdsAt(ULRPn,T).
\end{verbatim}}
\noindent
where all \code{holdsAt/2} terms share the same timestamp \code{T}, since the disjunction of conclusion ULRs \code{ULRC1}, ..., \code{ULRCm} holds immediately if all premise ULRs \code{ULRP1}, ..., \code{ULRPn} hold simultaneously at \code{T}.

\KALMR incorporates temporal information in queries also using \code{holdsAt/2}.
In this representation, the second argument of \code{holdsAt/2} is set to the highest value in the temporal domain extracted from the narrative.
For Example \ref{exmp:13}, a time-related query can be represented as
\code{holdsAt(ULRQ,3)?},
where \code{ULRQ} is the ULR of the query and \code{3} is the timestamp that exceeds all the explicitly given timestamps.

\section{\texorpdfstring\KALMR{} Evaluation}\label{sec:eval}
In this section, we assess the effectiveness of \KALMR-based knowledge authoring using two test suites, the clinical UTI guidelines (\cite{committee1999practice}) and the bAbI Tasks (\cite{weston2015towards}).

\subsection{Evaluation of Rule Authoring}
The UTI guidelines (\cite{committee1999practice}) is a set of therapeutic recommendations for the initial Urinary Tract Infection (UTI) in febrile infants and young children.
The original version in English was rewritten into the ACE 
CNL (\cite{shiffman2009writing}) for the assessment of ACE's expressiveness.
We rewrite the original English version into factual English, as shown in \ref{apdx:uti}.
This new version has a significant number of rules with disjunctive heads, as is common in the real-world medical domain.

The experimental results show that \KALMR is able to convert the UTI guidelines document into ULRs with 100\% accuracy.

\subsection{Evaluation of Authoring of Actions}
The 20 bAbI tasks (\cite{weston2015towards}) were designed to evaluate a system's capacity for natural language understanding, especially when it comes to actions.
They cover a range of aspects, such as 
moving objects (tasks 1-6),
positional reasoning (task 17), and 
path finding (task 19). 
Each task provides a set of training and test data, where each data point consists of a textual narrative, a question about the narrative, and the correct answer.
Fig.~\ref{fig:babi} in \ref{apdx:babi} presents 20 data points from 20 bAbI tasks respectively.
We used the test data for evaluation. Each task in the test data has 1,000 data points.

\begin{table}[htbp]
\caption{Result Comparisons}
\begin{tabular}{l|c|c|ccc}
\hline
        & STM  & ILA & \multicolumn{3}{c}{\KALMR}          \\ \hline
TASK   & Acc.  & Acc.                & \#I\&T & \#Rules & Acc.  \\ \hline
1 Single Supporting Fact       & 100  & 100                & 2          & 0              & 100  \\
2 Two Supporting Facts      & 99.79 & 100                & 4          & 1              & 100  \\
3 Three Supporting Facts      & 97.87 & 100                & 4          & 1              & 100  \\
4 Two Argument Relations      & 100  & 100                & 4          & 0              & 100  \\
5 Three Argument Relations      & 99.43 & 100                & 4          & 0              & 100 \\
6 Yes/No Questions      & 100  & 100                & 2          & 0              & 100  \\
7 Counting      & 99.19 & 100                & 4          & 0              & 100  \\
8 Lists/Sets      & 99.88 & 100                & 4          & 0              & 100  \\
9 Simple Negation      & 100  & 100                & 4          & 0              & 100  \\
10 Indefinite Knowledge     & 99.97 & 100                & 2          & 0              & 100    \\
11 Basic Coreference     & 99.99 & 100                & 2          & 0              & 100  \\
12 Conjunction     & 99.96  & 100                & 2          & 0              & 100  \\
13 Compound Coreference     & 99.99 & 100                & 2          & 0              & 93.1 \\
14 Time Reasoning     & 99.84  & 100                & 2          & 0              & 100  \\
15 Basic Deduction     & 100  & 100                & 0          & 1              & 100  \\
16 Basic Induction     & 99.71 & 93.6               & 2          & 1              & 93.6 \\
17 Positional Reasoning     & 98.82 & 100                & 8          & 20             & 100  \\
18 Size Reasoning     & 99.73 & 100                & 0          & 1              & 100  \\
19 Path Finding     & 97.94 & 100                & 12          & 4              & 100  \\
20 Agent's Motivations     & 100  & 100                & 5          & 6              & 100  \\ \hline
Average & 99.61 & 99.68              & 3.45        & 1.75            & 99.34\\ \hline
\end{tabular}
\label{tab:results}
\end{table}

The comparison systems in this evaluation include a state-of-the-art neural model on bAbI Tasks, STM (\cite{le2020self});
an approach based on inductive learning and logic programming (\cite{mitra2016addressing}) that we call LPA here;
and a recent sensation, ChatGPT.
The comparison with STM and ILA results are displayed in Table \ref{tab:results}, where ``\#I\&T" denotes the number of user-given initiation and termination statements (Definition \ref{def:2}) used to specify each particular task in \KALMR. 
The table shows that \KALMR achieves accuracy comparable to
STM and ILA.
%
%
%
ChatGPT has shown impressive ability to give correct answers
for some manually-entered bAbI tasks even though (we assume) it was not trained on that data set.
However, it quickly became clear that it has no robust semantic model behind its impressive performance and it makes many mistakes on bAbI Tasks. 
The recent (Jan 30, 2023) update of ChatGPT fixed some of the cases, while still not being able to handle slight perturbations of those cases.
Three such errors are shown in Table \ref{tab:chatgpt},
which highlights the need for authoring approaches, like \KALMR, which are based on robust semantic models.

\begin{table}[hbtp]
\caption{ChatGPT Error Cases}
\begin{tabular}{l|l|l}
\hline
\multicolumn{1}{c|}{\textbf{Task 2}}             & \multicolumn{1}{c|}{\textbf{Task 17}}              & \multicolumn{1}{c}{\textbf{Task 19}}      \\
\multicolumn{1}{c|}{\textbf{2 Supporting Facts}} & \multicolumn{1}{c|}{\textbf{Positional Reasoning}} & \multicolumn{1}{c}{\textbf{Path Finding}} \\ \hline
Mary went to the kitchen.                    & The red square is below                                 & The garden is west of the hallway.        \\
Mary got the apple.                              & \hspace{7mm}the blue square.                             & The kitchen is west of the garden.        \\
Mary got the ball.                               & The red square is left of                         & The garden is north       \\
Mary got the book.                               & \hspace{7mm}the pink rectangle.                        & \hspace{7mm}of the bathroom.      \\
Mary went to the bedroom.                        &                                                    & The bedroom is east        \\
Mary went to the garden.                    &                                                    & \hspace{7mm}of the bathroom.                                          \\
Mary dropped the book.                           &                                                    & The hallway is west of the office.                                          \\ \hline
Q: Where is the apple?                          & Q: Is the blue square below                        & Q: How do you go from the                 \\
                                                 & \hspace{4.6mm}the pink rectangle?                                & \hspace{4.6mm}bathroom to the hallway?                  \\ \hline
ChatGPT: ... not specified                      & ChatGPT: ... not specified                         & ChatGPT: ... east..., ... south...        \\
Correct: \hspace{3.6mm}garden                                   & Correct: \hspace{3.6mm}no                                        & Correct: \hspace{3.6mm}east, north                      \\ \hline
\end{tabular}
\label{tab:chatgpt}
\end{table}

As to \KALMR, it does not achieve 100\% correctness on Tasks 13 (Compound Coreference) and 16 (Basic Induction).
In Task 13, the quality of \KALMR's coreference resolution is entirely dependent on the output of neuralcoref, the coreference resolver we used. As this technology improves, so will \KALMR.
Task 16 requires the use of the induction principles adopted by bAbI tasks, some of which are questionable.
For instance,
in Case 2 of Table~\ref{tab:errors}, the color is determined by the maximum frequency of that type, whereas in Case 3, the latest evidence determines the color.
Both of these principles are too simplistic and, worse, contradict each other.


\begin{table}[hbtp]
\caption{\KALMR Error Cases}
\begin{tabular}{l|ll}
\hline
\multicolumn{1}{c|}{\textbf{Case1}}                       & \multicolumn{1}{c|}{\textbf{Case 2}}        & \multicolumn{1}{c}{\textbf{Case 3}} \\ \hline
\multicolumn{1}{c|}{\textbf{Task 13 Compound Coreferences}} & \multicolumn{2}{c}{\textbf{Task 16 Basic Induction}}                              \\ \hline
Mary and Sandra went                                      & \multicolumn{1}{l|}{Brian is a swan.}       & Berhnard is a rhino.                \\
\hspace{7.2mm}back to the bedroom.                                      & \multicolumn{1}{l|}{Greg is a swan.}        & Brian is a rhino.                   \\
Then they moved to the kitchen.                            & \multicolumn{1}{l|}{Julius is a swan.}      & Bernhard is white.                  \\
Sandra and Daniel went                                    & \multicolumn{1}{l|}{Greg is gray.}          & Brian is white.                     \\
\hspace{7.2mm}back to the bathroom.                                     & \multicolumn{1}{l|}{Julius is gray.}        & Lily is a lion.                     \\
Then they went to the office.                             & \multicolumn{1}{l|}{Bernhard is a lion.}    & Lily is yellow.                     \\
                                                          & \multicolumn{1}{l|}{Lily is a swan.}        & Greg is a rhino.                    \\
                                                          & \multicolumn{1}{l|}{Berhnard is green.}     & Greg is green.                      \\
                                                          & \multicolumn{1}{l|}{Brian is white.}        & Julius is a rhino.                  \\ \hline
Q: Where is Daniel?                                       & \multicolumn{1}{l|}{Q: What color is Lily?} & Q: What color is Julius?            \\ \hline
\KALMR: bathroom                                            & \multicolumn{1}{l|}{\KALMR: gray, white}      & \KALMR: green, white                  \\
bAbI Correct: office                                           & \multicolumn{1}{l|}{bAbI Correct: gray}          & bAbI Correct: green                      \\ \hline
\end{tabular}
\label{tab:errors}
\end{table}

\section{Conclusion and Future Work}\label{sec:conclusion}
The \KALMF system (\cite{wang2022knowledge}) was designed to address the limitations of KALM (\cite{gao2018high}) in terms of expressive power and the costs of the actual authoring of knowledge by human domain experts.
KALM did not support authoring of rules and actions, and it required abiding a hard-to-learn grammar of the ACE CNL.
In this paper, we introduced \KALMR, an NLP system that extends \KALMF to authoring of rules and actions by tackling a slew of problems.
The evaluation results show that \KALMR achieves 100\% accuracy on authoring rules, and 99.34\% accuracy on authoring and reasoning with actions, demonstrating the effectiveness of \KALMR at capturing knowledge via facts, actions, rules, and queries.
In future work, we plan to add non-monotonic extensions of factual English to support defeasible reasoning (\cite{WanGKFL09}),
a more natural way of human reasoning in real life, 
where conclusions are derived from default assumptions, but some conclusions may be retracted when the addition of new knowledge violates these assumptions.

\bibliographystyle{tlplike}
\bibliography{main}

\newpage
\appendix

\section{Complete UTI Guidelines in Factual English}\label{apdx:uti}
\textbf{Background Rules}
\begin{enumerate}
    \item \textbf{If} \textit{\$doctor} administers \textit{\$therapy} for \textit{\$patient}, \textbf{then} \textit{\$patient} undergoes \textit{\$therapy} from \textit{\$doctor}.
    \item \textbf{If} \textit{\$patient} undergoes \textit{\$therapy} from \textit{\$doctor}, \textbf{then} \textit{\$patient}'s \textit{\$therapy} from \textit{\$doctor} is completed, or not completed.
    \item \textbf{If} \textit{\$doctor} performs \textit{\$imaging\_study} for \textit{\$patient}, \textbf{then} \textit{\$patient}'s \textit{\$imaging\_study} from \textit{\$doctor} is completed, or not completed.
\end{enumerate}

\noindent
\textbf{Recommendation 1}
\begin{enumerate}
    \item \textbf{If} \textit{\$doctor}'s \textit{\$patient} is a young child and has an unexplained fever, \textbf{then} \textit{\$doctor} considers UTI for \textit{\$patient}.
\end{enumerate}

\noindent
\textbf{Recommendation 2}
\begin{enumerate}
    \item \textbf{If} \textit{\$doctor}'s \textit{\$patient} is a young child and has an unexplained fever, \textbf{then} \textit{\$doctor} assesses \textit{\$patient}'s degree of toxicity.
    \item \textbf{If} \textit{\$doctor}'s \textit{\$patient} is a young child and has an unexplained fever, \textbf{then} \textit{\$doctor} assesses \textit{\$patient}'s degree of dehydration.
    \item \textbf{If} \textit{\$doctor}'s \textit{\$patient} is a young child and has an unexplained fever, \textbf{then} \textit{\$doctor} assesses \textit{\$patient}'s ability to retain oral intake.
    \item \textbf{If} \textit{\$doctor} assesses \textit{\$patient}'s ability to retain oral intake, \textbf{then} \textit{\$doctor}'s \textit{\$patient} is or is not able to retain oral intake.
\end{enumerate}

\noindent
\textbf{Recommendation 3}
\begin{enumerate}
    \item \textbf{If} \textit{\$doctor}'s \textit{\$patient} is a young child and has an unexplained fever, and \textit{\$doctor}'s \textit{\$patient} is sufficiently ill, \textbf{then} \textit{\$doctor}  analyzes the culture of \textit{\$patient}'s urine specimen obtained by SPA or transurethral catheterization.
\end{enumerate}

\noindent
\textbf{Recommendation 4}
\begin{enumerate}
    \item \textbf{If} \textit{\$doctor}'s \textit{\$patient} is a young child and has an unexplained fever, and \textit{\$doctor}'s \textit{\$patient} is not sufficiently ill, \textbf{then} \textit{\$doctor}  analyzes the culture of \textit{\$patient}'s urine specimen obtained by SPA, transurethral catheterization, or a convenient method.
    \item \textbf{If} \textit{\$doctor}'s \textit{\$patient} is a young child and has an unexplained fever, \textit{\$doctor}'s \textit{\$patient} is not sufficiently ill, \textit{\$doctor} analyzes the culture of \textit{\$patient}'s urine specimen obtained by a convenient method, and the analysis of \textit{\$patient}'s culture of a urine specimen suggests UTI, \textbf{then} \textit{\$doctor}  analyzes \textit{\$patient}'s culture of a urine specimen obtained by SPA or transurethral catheterization.
    \item \textbf{If} \textit{\$doctor} analyzes the culture of \textit{\$patient}'s urine specimen obtained by SPA or transurethral catheterization, \textbf{then} \textit{\$doctor}'s analysis of \textit{\$patient}'s culture confirms UTI or excludes UTI.
    \item \textbf{If} \textit{\$doctor} analyzes the culture of \textit{\$patient}'s urine specimen obtained by a convenient method, \textbf{then} \textit{\$doctor}'s analysis of \textit{\$patient}'s culture suggests UTI or does not suggest UTI.
    \item \textbf{If} \textit{\$doctor}'s analysis of the culture of \textit{\$patient}'s urine specimen confirms UTI, \textbf{then} \textit{\$doctor}'s \textit{\$patient} has UTI.
    \item \textbf{If} \textit{\$doctor}'s analysis of the culture of \textit{\$patient}'s urine specimen excludes UTI, \textbf{then} \textit{\$doctor}'s \textit{\$patient} does not have UTI.
\end{enumerate}

\noindent
\textbf{Recommendation 5} is integrated into 3 and 4.

\noindent
\textbf{Recommendation 6}
\begin{enumerate}
    \item \textbf{If} \textit{\$doctor}'s \textit{\$patient} is a young child and has an unexplained fever, and \textit{\$doctor}'s \textit{\$patient} is toxic, \textbf{then} \textit{\$doctor} administers an antimicrobial therapy for \textit{\$patient}.
    \item \textbf{If} \textit{\$doctor}'s \textit{\$patient} is a young child and has an unexplained fever, and \textit{\$doctor}'s \textit{\$patient} is toxic, \textbf{then} \textit{\$doctor} considers hospitalization for \textit{\$patient}.
    \item \textbf{If} \textit{\$doctor}'s \textit{\$patient} is a young child and has an unexplained fever, and \textit{\$doctor}'s \textit{\$patient} is dehydrated, \textbf{then} \textit{\$doctor} administers an antimicrobial therapy for \textit{\$patient}.
    \item \textbf{If} \textit{\$doctor}'s \textit{\$patient} is a young child and has an unexplained fever, and \textit{\$doctor}'s \textit{\$patient} is dehydrated, \textbf{then} \textit{\$doctor} considers hospitalization for \textit{\$patient}.
    \item \textbf{If} \textit{\$doctor}'s \textit{\$patient} is a young child and has an unexplained fever, and \textit{\$doctor}'s \textit{\$patient} is not able to retain oral intake, \textbf{then} \textit{\$doctor} administers an antimicrobial therapy for \textit{\$patient}.
    \item \textbf{If} \textit{\$doctor}'s \textit{\$patient} is a young child and has an unexplained fever, and \textit{\$doctor}'s \textit{\$patient} is not able to retain oral intake, \textbf{then} \textit{\$doctor} considers hospitalization for \textit{\$patient}.
\end{enumerate}

\noindent
\textbf{Recommendation 7}
\begin{enumerate}
    \item \textbf{If} \textit{\$doctor}'s \textit{\$patient} is a young child, and \textit{\$doctor}'s analysis of the culture of \textit{\$patient}'s urine specimen confirms UTI, \textbf{then} \textit{\$doctor}  administers a parenteral or oral antimicrobial therapy for \textit{\$patient}.
\end{enumerate}

\noindent
\textbf{Recommendation 8}
\begin{enumerate}
    \item \textbf{If} \textit{\$doctor}'s \textit{\$patient} is a young child and has UTI, \textit{\$patient} undergoes an antimicrobial therapy from \textit{\$doctor} for 2 days, and \textit{\$doctor}'s \textit{\$patient} does not show the expected response of the antimicrobial therapy, \textbf{then} \textit{\$doctor}  reevaluates \textit{\$patient} and analyze the culture of \textit{\$patient}'s second urine specimen.
    \item \textbf{If} \textit{\$doctor}'s \textit{\$patient} is a young child and has UTI, and \textit{\$patient} undergoes an antimicrobial therapy from \textit{\$doctor} for 2 days, \textbf{then} \textit{\$doctor}'s \textit{\$patient} shows or does not show the expected response of the antimicrobial therapy.
\end{enumerate}

\noindent
\textbf{Recommendation 9}
\begin{enumerate}
    \item \textbf{If} \textit{\$doctor}'s \textit{\$patient} is a young child and has UTI, \textbf{then} \textit{\$doctor} administers an oral antimicrobial therapy that lasts at least 7 days for \textit{\$patient}.
    \item \textbf{If} \textit{\$doctor}'s \textit{\$patient} is a young child and has UTI, \textbf{then} \textit{\$doctor} administers an oral antimicrobial therapy that lasts at most 14 days for \textit{\$patient}.
\end{enumerate}

\noindent
\textbf{Recommendation 10}
\begin{enumerate}
    \item \textbf{If} \textit{\$doctor}'s \textit{\$patient} is a young child and has UTI, the antimicrobial therapy of \textit{\$doctor}'s \textit{\$patient} is completed, and the imaging study of \textit{\$doctor}'s \textit{\$patient} is not completed, \textbf{then} \textit{\$doctor} administers \textit{\$patient} a therapeutically or prophylactically dosed antimicrobial.
\end{enumerate}

\noindent
\textbf{Recommendation 11}
\begin{enumerate}
    \item \textbf{If} \textit{\$doctor}'s \textit{\$patient} is a young child and has UTI, \textit{\$patient} undergoes an antimicrobial therapy for 2 days from \textit{\$doctor}, and \textit{\$doctor}'s \textit{\$patient} does not show the expected response of the antimicrobial therapy, \textbf{then} \textit{\$doctor} performs ultrasonography promptly for \textit{\$patient}.
    \item \textbf{If} \textit{\$doctor}'s \textit{\$patient} is a young child and has UTI, \textit{\$patient} undergoes an antimicrobial therapy for 2 days from \textit{\$doctor}, and \textit{\$doctor}'s \textit{\$patient} does not show the expected response of the antimicrobial therapy, \textbf{then} \textit{\$doctor} performs VCUG or RNC for \textit{\$patient}.
    \item \textbf{If} \textit{\$doctor}'s \textit{\$patient} is a young child and has UTI, \textit{\$patient} undergoes an antimicrobial therapy for 2 days from \textit{\$doctor}, and \textit{\$doctor}'s \textit{\$patient} shows the expected response of the antimicrobial therapy, \textbf{then} \textit{\$doctor} performs VCUG or RNC for \textit{\$patient}.
    \item \textbf{If} \textit{\$doctor}'s \textit{\$patient} is a young child and has UTI, \textit{\$patient} undergoes an antimicrobial therapy for 2 days from \textit{\$doctor}, and \textit{\$doctor}'s \textit{\$patient} shows the expected response of the antimicrobial therapy, \textbf{then} \textit{\$doctor} performs VCUG or RNC for \textit{\$patient}.
\end{enumerate}


\newpage
\section{Sample Narratives and Questions from 20 bAbI Tasks}\label{apdx:babi}
\begin{figure}[htbp!]
    \centering
    \includegraphics[scale=0.33]{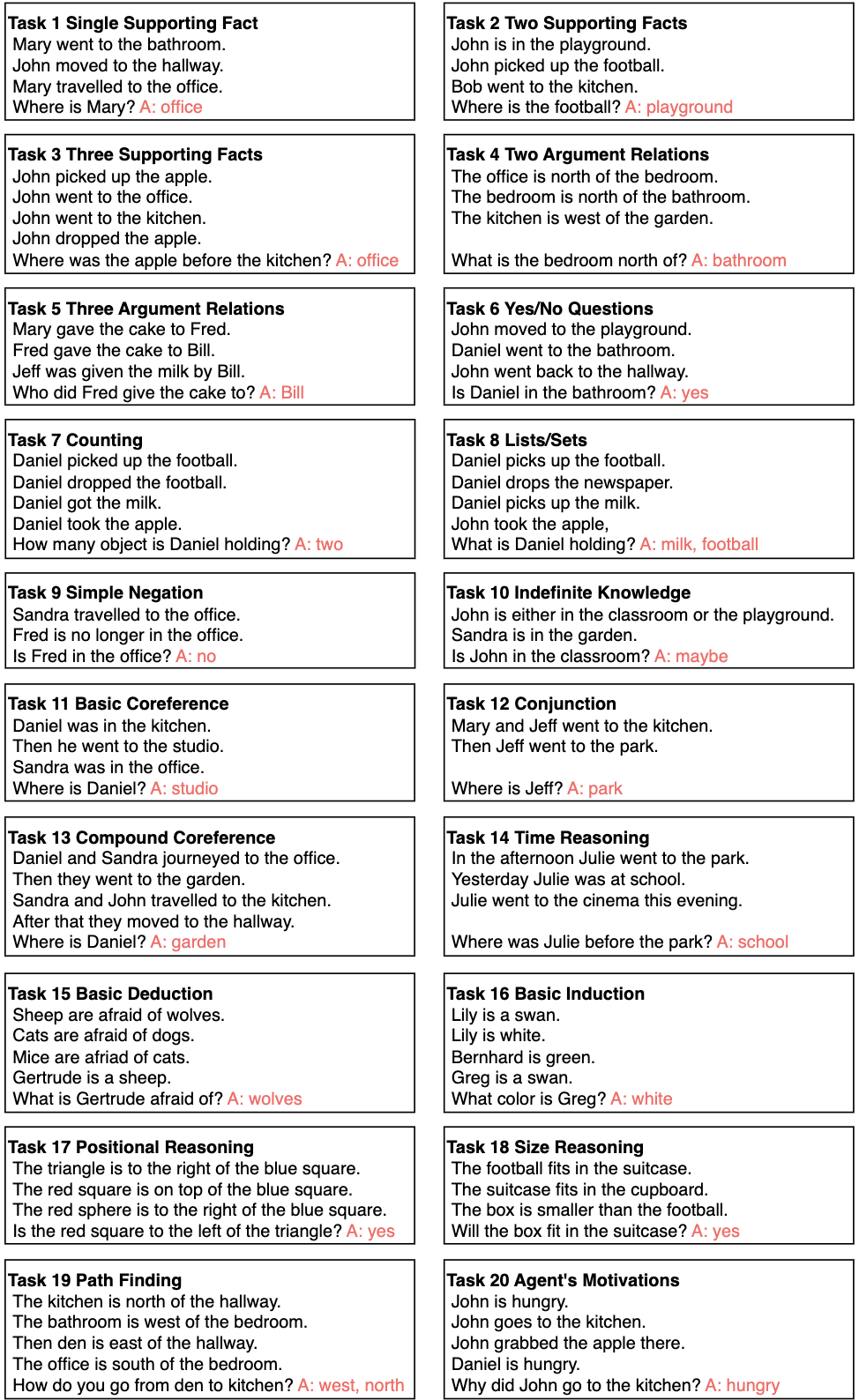}
    \caption{Data points from the 20 bAbI Tasks}
    \label{fig:babi}
\end{figure}

\end{document}